\documentclass[runningheads]{llncs}
\usepackage{graphicx}
\usepackage[numbers,sort&compress]{natbib}

\usepackage{tikz}
\usepackage{comment}
\usepackage{amsmath,amssymb} 
\usepackage{color}
\usepackage{float}

\newcommand\blfootnote[1]{%
  \begingroup
  \renewcommand\thefootnote{}\footnote{#1}%
  \addtocounter{footnote}{-1}%
  \endgroup
}

\usepackage{hyperref}
\hypersetup{
    colorlinks=true,
    linkcolor=blue,
}
\usepackage{subcaption}

\begin{document}
\pagestyle{headings}
\mainmatter
\def\ECCVSubNumber{100}  

\title{Deep Learning for Cornea Microscopy Blind Deblurring} 


\titlerunning{Deep Learning for Cornea Microscopy Blind Deblurring}
%
\author{Toussain Cardot \and
Pilar Marxer \and
Ivan Snozzi}
\authorrunning{Toussain Cardot, Pilar Marxer \and Ivan Snozzi}
%
\institute{Ecole Polytechnique Fédérale de Lausanne (EPFL), Switzerland}
\maketitle
\blfootnote{Supervised by Majed El Helou}
\begin{abstract}
The goal of this project is to build a deep-learning solution that deblurs cornea scans, used for medical examination. The spherical shape of the eye prevents ophtamologist to have a complete sharp image. Provided with a stack of cornea from confocal images, our approach is to build a model that performs an upscaling of the images using SR (Super Resolution) Network. We used the net of \cite{arXiv:2003.05961} and adjust it to our problem. The task of image debluring is well documented in the literature, and powerful solutions have been developed. However microscopy images still represent a challenge for deep learning.
\dots
\keywords{Out-of-focus blur, Deblurring, Confocal Microscopy Images, Neural network }
\end{abstract}
\newpage
\section{Introduction}

Image deblurring is a fundamental problem in the field of image processing and computer vision. The aim being to recover a sharp image from its blurred source, this problem is highly ill-posed and multiple strategies can be deployed to solve it. However, the underlying assumption stays the same: the blur is modeled by an unknown blur kernel \textit{k}, also known as the \textit{point-spread-function}, which convoluted with our sharp image (with addition of white noise) produces a blurry output image. In our specific case, we handle images coming from an eye scan, the nature of the blur thus comes from the part which are out of focus. Our solution takes as input single-focus cornea scans, with a certain depth range that is originally in focus, and extends the depth of field by deblurring as much as possible around that range. Solving this problem could be useful for the biomedical imaging community since cell counting is utilized to detect diseases correlated with cell density in the cornea.\\
The problem of deblurring can be divided into two categories: blind deblurring and non-blind deblurring. In the latter the blur kernel is known and the solution boils down to a simple deconvolution to retreive the sharp image. For non-blind deblurring we do not know the blur kernel, and the problem is thus more challenging, and applies to most of the applications involving natural images including ours. Also the nature of the blur requires different approaches, in our case we handle an out-of-focus blur generated by the 3 dimensional shape of the eye thus our method aims to reclaim depth from 2D images, which is analogous to estimating the third physical dimension lost during the imaging process. This contrasts with motion blur (from the camera or the object), which is much more explored in the literature. Multiple deep-learning solutions have been proposed to solve it, but those do not fit well our problem since the underlying physics are inherently different. Here are the main papers we found meaningful to cite, including full convolutional networks \cite{8099518}, generative adversarial networks \cite{8578952} and solutions using prior learning \cite{arXiv:1503.00593}.
Traditional approaches for out-of-focus deblurring handle the problem by estimating the blur kernel (and later the inverse of kernel of the estimation) using underlying characteristics to create a model, and along with different image priors, the solution space can be regularized. This in an attempt to simplify the problem and shift it to a non-blind deblurring one. Ideas of  methods are described by \cite{5559018}. 
 However, for most real-images, this does not fit well the blur patterns being more complicated and coming from various distortions, simple kernel deconvolution cannot model several challenging cases such as occluded regions or depth variations. Furthermore, kernel estimation process
is subtle and sensitive to noise and saturation, unless the blur
model is carefully designed. To find a deblurring model for those, kernel-free learning-based methods are preferred. Here are  also different approaches that have been explored as cited for motion blurring. 
\\
One of latest solution implemented by Ye et al. \cite{8963625} describes a Scale-Iterative Upscaling Network for natural image deblurring. Their solution adapts to different type of images (face, text, nature) whose blur may be caused by various combined distortions, this is innovative since a lot of the methods focus on one type of images and in our personal documentation it was hard to find something similar to our research (i.e. microscopy images). They exploit the idea that, to preserve and recover details upscaling, layers can be used. In most previous implementations this was done using simple upsampling layers but this is not enough to achieve satisfying results. To solve this problem, they use the structure of SR to replace the upsampling layer, aiming to reconstruct a high-resolution (HR) image from its low-resolution (LR) counterpart so that the high frequency information can be well mapped to the next scale level. They use Residual Dense Network for details restoration, as it makes full use of both global and local hierarchical features. Along with some other interesting specifications which we will discuss further in the literature, this solution beats the traditional and learning based solutions developed so far. Our supervisor suggested to use a similar approach before reading this paper, their approach is much more complex and shows interesting insights so it is meaningful to cite it here. \\
In our final solution we reused the super-resolution structure of the network implemented for \cite{arXiv:2003.05961} and adjusted it to our problem. It follows the state of art in the field of deblurring but our solution focuses on confocal microscopy images which is different from the work done by Ye et al., furthermore we do not have a large dataset at our disposal, in fact we had less than 10 ground truth images. This is also a major drawback knowing that big data sets are generally required to obtain good results with deep learning approaches. Furthermore microscopy images tend to be similar in general. This lack of  diversity can also impact the network's training and performances. So the way we tackled the data preprocessing was crucial, we had to come up with methods to augment our training dataset and simulate blur kernels similar to the real kernels of our original images. The goal of the project was to understand the problem in depth and implement a solution following the state-of-the-art, while keeping in mind that the final application is cell counting. In a first stage, we had to get familiar with the available deep learning solutions for out-of-focus deblurring and understanding the physical problem behind as well as the involved implications for deep learning. Choosing which one would best fit our problem did not happen in one single trial. We explored multiple approaches, starting with a classification task, since those are the most familiar with us. Classification tasks, imply the learning of a prior at that stage the difference between kernel-estimation problems with MAP priors and kernel-free learning methods became more clear. Especially, this paper inspired us in the early stages \cite{Authors14b}, because they were treating microscopy images similar to us, but in the end their implementation goal is quite different than ours. Kernel-free estimation methods seemed to be the way to go as they do not need much post processing. Furthermore the recent articles are using this kind of methods and have better results in general for the reasons mentioned earlier. Also kernel-free methods would be the "nice to have" solution, meaning that the image is simply fed to the network which returns it sharp. Here we face a dilemma, writing our personal network from scratch would impede the finding of a good solution. We first studied the implementation of \cite{art4}, their code was developed in \textit{Matlab} back in 2014, deep nets had to be implemented from scratch without pre-implemented libraries, which make them  not convenient to reuse. We then continued to look for other nets, and  our supervisor finally suggested to reuse the code developed for a W2S: A Joint Denoising and
Super-Resolution Dataset, which we adjusted to our needs as good as we could. In parrallel to the quest of our deep-learning pipeline, we needed to generate training data and adjust the original cornea stack images to generate test data. Details on the ground truth images and the different blur simulation and data augmentations will be explained in details in the implementations section.\\
In the following sections we will go over the literature related to our problem, the details of our implementation including data generation, training and post processing, and finally we will discuss our results and encountered issues.
\section{Literature}
In this section is dedicated to the review of the literature related to learning-based image deblurring. We will highlight different papers that are useful to understand the algorithms used for our solution. There will be different approaches for kernel estimation using neural network (also known as image prior learning), deconvolution methods, depth estimation to deblur images, and finally RRDB networks and their use for different tasks including blur removal.
\subsection{Prior Learning}
Prior learning is a classical approach in the family of deblurring solutions, so it does make sense to give additional insights on the methodology by showing an example of work here. Sun et al. \cite{arXiv:1503.00593} proposed a sequential deblurring approach where they first generated pairs of blurry and sharp patches with candidate blur kernels (motion blur kernels here, but works in the same manner for out-of-focus blur kernel), then they trained a classification CNN to measure the likelihood of a specific blur kernel on a local patch and by smoothly varying blur kernel (the predicted kernel have to follow a smooth motion over the whole image) their final result is obtained by optimizing an energy model that is composed of the CNN likelihoods and smoothness priors. In the end the sharp image is obtained by performing deconvolution on the patches with their assigned blur-kernel and a conventional optimization method for good results. We chose not to follow this type of approach, recent papers show end-to-end deconvolution networks which are more powerful and the problem is narrowed down to one network which is more ideal if one thinks of applications in the industry.
\subsection{Fourier-Domain Optimization for Image Processing}
The work of \cite{arXiv:1809.04187} describes a method for deconvolution and how optimization can be done, it is addressed to a broad audience with minimal knowledge in convolution and image optimization. When the deblurring is done using a kernel estimation it is worth spending time in optimizing the deconvolution algorithm. The mathematical background along with the practical implementations are described and an open-source demo is made available. A key point that the authors take adavantage of is that a time/spacedomain convolution corresponds to a multiplication in the Fourier domain. Knowing that our problem consists of \textit{blurr}=k*\textit{sharp} + n (where k is the point-spread-function and n some white noise) it is a crutial task to implement an optimized deconvolution for good and fast image reconstruction. In addition, images are generally large matrices thus handling them in the Fourier Domain is challenging.   
\subsection{Depth Estimation and Blur Removal from a
Single Out-of-focus Image}
Out-of-focus blur varies with the depth of the images. Reconstructing depth from 2D images is analogous to estimating the third physical dimension lost during the imaging process.Their method is powerful because it can reconstruct an all-focus image and achieve synthetic re-focusing, all from a single image. The paper of \cite{art3} takes advantage of this aspect, and implements a deep net that identifies the different depths of the image pixels. Once these are computed it applies a deconvolution on each pixel with a different kernel for each depth. This solution is nice because it follows a reasoning that is in accordance with the physics of the problem of the project, even though it is not one of the latest. 
\subsection{Residual Dense Network for Image Super-Resolution \cite{DBLP:journals/corr/abs-1802-08797}}
The goal of single image Super-Resolution (SISR) is to retrieve an high resolution image from its degraded version. This is exactly our task, since the data set of cornea scans we have has been degraded by the blur. 
\\
To achieve their goal they utilize a Residual Dense Network (RDN) which is not too different from the one we have in our implementation. The RDN they propose consist in a series of RDB, where each block consist of densely connected layers with local feature fusions and local residual learning.
\begin{figure}
    \centering
    \includegraphics[scale =0.3]{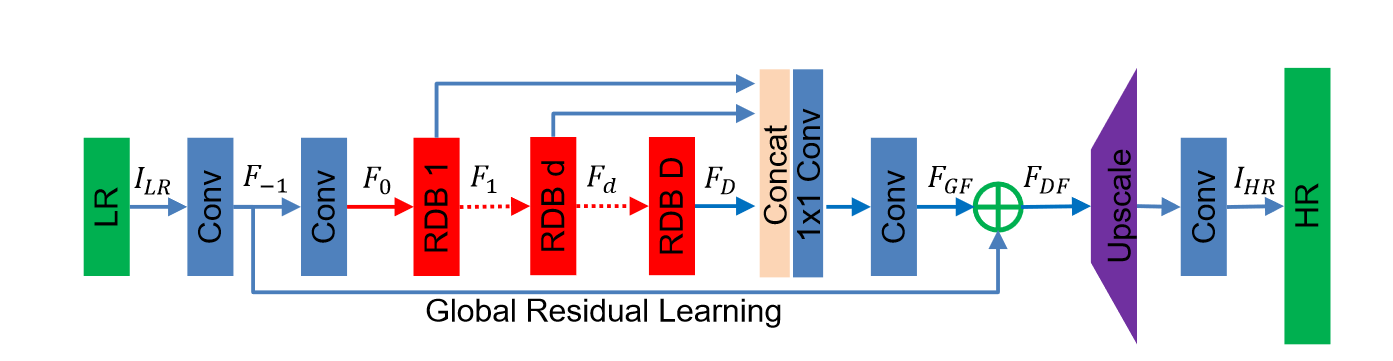}
    \caption{RDN source: \cite{DBLP:journals/corr/abs-1802-08797}}
    \label{fig:5}
\end{figure}
\subsection{W2S: A Joint Denoising and
Super-Resolution Dataset}
Part of the implementation in our final solution is derived from \cite{arXiv:2003.05961}, even though their ultimate goal was different. They describe the aquisition of a dataset for the purpose of simultaneous denoising and super-resolution. It is relevant to mention that their dataset is generated out of widefield microscope images, which are in the same class as ours. Still, our images come from confocal microscopy so there are major differences in the artifacts. Furthermore their dataset consisted of about 120 images, which in comparison to ours is much bigger and thus the results obtained with their neural net will differ to ours. We retained some of their data preprocessing details, Image alignement techniques are described, they use brute force matching based on the Hamming distance, with ORB. For our project image alignment was also necessary when we create ground truth images from the eye scans, more details in the implementation part.
\subsection{Scale-Iterative Upscaling Network for
Image Deblurring}
The paper of Ye et al. \cite{8963625} use similar tools to us, as already mentioned in the introduction but with a different approach. The network starts from a down-sampled scale and
works in an iterative way. For each iteration, the output is
upscaled until a full resolution image is restored. Thus the benefit to use an RDN-based super-resolution architecture for a better preserving of the features from previous level. In addition a curriculum learning strategy is applied. Curriculum learning means learning by gradually increasing
the difficulty of the task. In this case the difficulty of the sub-tasks (from iterations) increases gradually,
because blur magnitude decreases when the blurry image is
down-sampled and their network thus converges faster. 
\begin{figure}
    \centering
    \includegraphics[scale = 0.7]{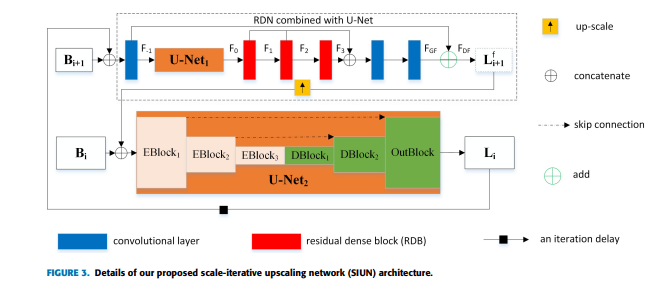}
    \caption{RDN source: \cite{8963625}}
    \label{fig:5}
\end{figure}
\section{Implementation}
\subsection{Data Exploration}
The data we had at our disposal was composed of 21 single-focus RGB cornea scans of 4 different regions of the eye with size 2560x1920 pixels. We were not provided with the ground truth of the 4 regions. We decided to convert our data to graycale images, since most of the shape's information is not contained in the colors and because of performance reasons.
\begin{figure}
    \centering
    \includegraphics[scale = 0.40]{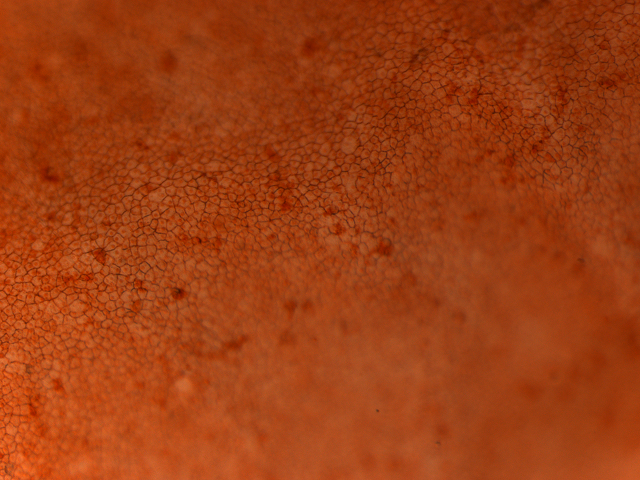}
    \caption{Cornea scan example from the original set}
    \label{fig:1}
\end{figure}
\subsubsection{Training data}
We were given a data set of 4 regions of the cornea photographed at different focus points. The focus were made such that one full sharp image of one region could be made by stacking the pictures. There were roughly 5 pictures by region for a total of 21 color pictures of size 1920x2560. As the neural network works on patches of 64x64 with a stride of 32 to blend the limits of the patches, the structure of the dataset is simply two sets of patches: one containing sharp patches (in focus) and another one with patches corresponding to the same region but blurred.  The set at our disposal is really small, especially for deep learning and one main concern is to augment our data. \\
The first step of the creation of such sets is gathering as many sharp patches as possible. Our first idea was to cut the images into patches and manually select the sharp ones. To achieve this, we first assembled the full stack (the image totally in focus) by combining manually images with different focus in \textit{Photoshop}. We used this method to obtain rapidly some training data but aligning the images manually is not precise and hardly reproducible. Therefore, we later took the realigned stacks from the other team working on the same deblurring project, but with traditional methods. \\
From the four full stacks, we generated 40 images by applying horizontal and vertical flips as well as rotations of 90, 180 and 270 degrees. Finally, the 40 images are cut into patches that will form the ground truth set. The next step is to make fake blurry patches. As the intensity of the blur varies over the image, we use several kernels of different intensity to blur the images and simulate an increase in depth. To begin, we blurred the patches with Gaussian kernels. The first set of blurred image was made with 50 kernels with standard deviations from 1 to 25, the kernels were applied to 4 images to obtain 200 blurred images in total. Once the images blurred, they are cut into patches, each i-th patch in the blurred set corresponds to the same region as the i-th one of the ground truth set. \\
Training the network with Gaussian blurs revealed to be successful (see results section) but incorrect from a modeling point of view as the out-of-focus blur is not the same as Gaussian blur. To achieve a better modeling and thus a training set closer to real blurred images, we re-created a training set using this time kernels estimated from the original blurry images (see kernel estimation section). We created a second data set using 20 estimated kernels applied to the 40 sharp images, reaching a total of 800 blurred images.
\subsection{Kernel estimation}
Traditionally, kernel estimation in deblurring is used to compute the inverse kernel of a blur to reconstitute a sharp image. In this project, kernel estimation is only used to create new data for the training of the neural network. We apply kernels that approach the real blur of the image to the full stacks, as if the full stack was photographed at different focus.\\
The kernel is estimated by comparing two images of one zone of the cornea: one in focus and the other out of focus. First, the images have to be aligned; we used the image registration tool from the Matlab Image Processing Toolbox. It takes two images as input: a fixed image and a moving image. It outputs the transform to apply on the moving image for it to be aligned to the fixed image This tool has several methods to detect features such as phase correlation, MSER and SURF. In our case, SURF had the best performances for finding matching features between the two images. SURF (Speeded Up Robust Features) is a descriptor algorithm partially based on SIFT (Scale-Invariant Feature Transform). Once enough similar features are found between the images, the blurred one is registered on the sharp. We applied this method to align several images partially in focus to the full stack. \\
To obtain a batch of different kernels, we cut into patches one full stack and one image partially in focus, we then applied the kernel estimation on each pair patches. This algorithm (taken from \cite{arXiv:2003.05961}) need to know the size of the kernel to estimate. Therefore we estimated, for each pair of patch, kernels from size 3x3 to 35x35 and selected the one that minimizes the most the MSE between the two patches. We also applied applied a denoiser ($denoise\_wavelet$ from Skimage) to blurry image before the kernel estimate, which reveal to reduce the MSE between the sharp image and the blurry (not denoised) image. \\
\begin{figure}[H]
    \centering
    \includegraphics[scale=0.2]{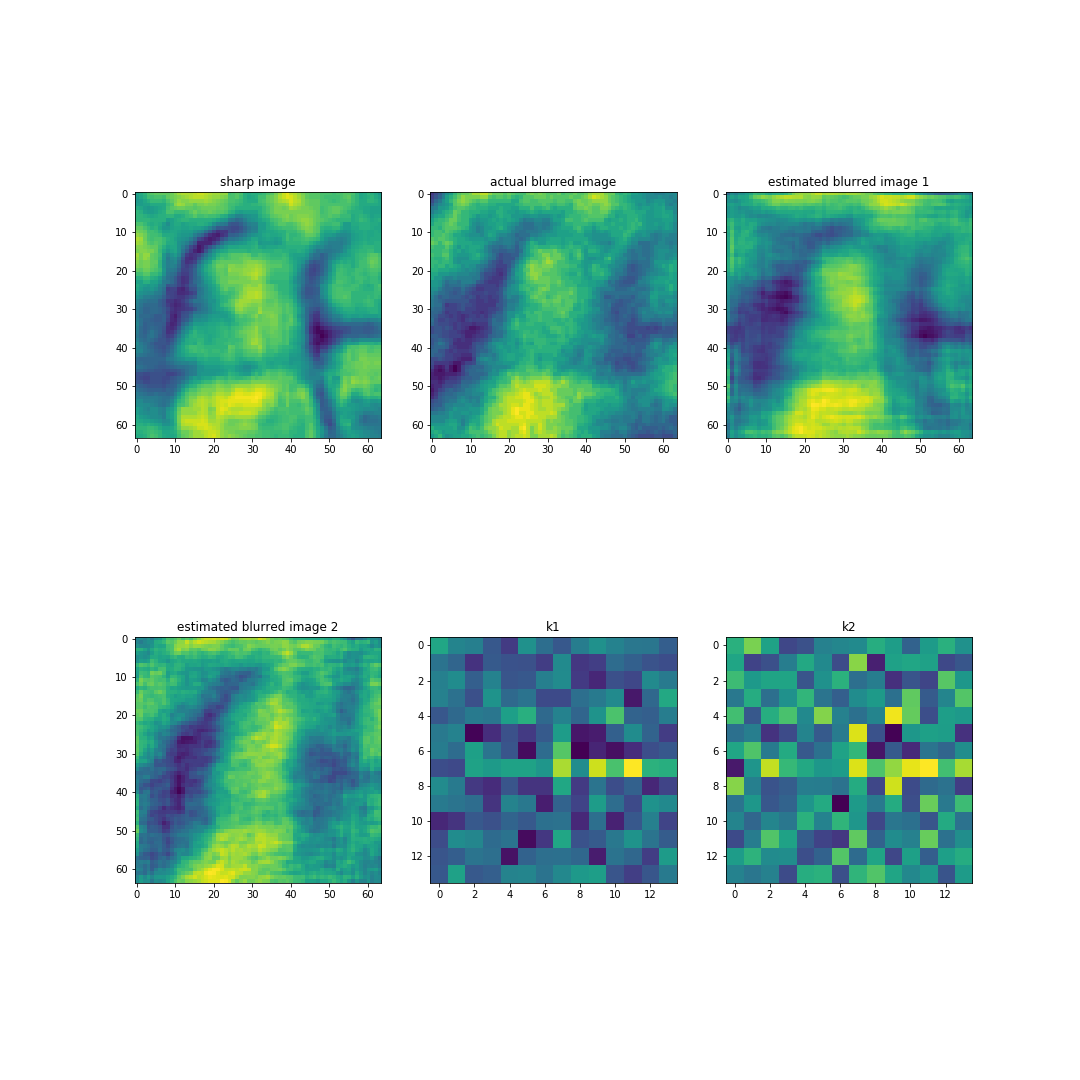}
    \caption{Comparison of kernel estimation with and without denoiser: k1 is estimated with a denoiser and k2 without. }
    \label{denoiserKernel}
\end{figure}
 The kernels for each pair of patches estimated, we then need to select the best ones to augment our training data. We suppose that small kernels are better estimated that big ones and thus just selecting the kernels that minimize the MSE would only give us small kernels. Before selecting the kernels, we put them in 3 categories: kernel with size from 3x3 to 10x10, from 11x11 to 20x20 and kernels of size higher than 20x20. We then selected the 5 best kernels for each of the two last categories. We ignored kernels of size less than 10x10 as they do not add a lot of blur to the patches and thus are not very useful to train the network.\\
We aligned two partially blurred images to one full stack and selected the best estimated kernels for, which gives us 20 kernels to blur the training data.
\begin{figure}[H]
    \centering
    \includegraphics[scale = 0.5]{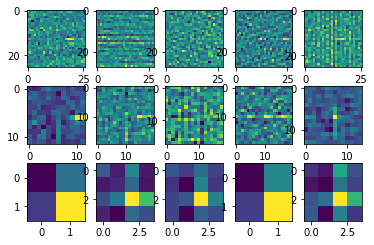}
    \caption{the 5 best kernel estimation for the 3 categories}
    \label{differentKernels}
\end{figure}
\subsection{The RRDBNet}
The network we utilized is a modified version of the RRDBnet from  \href{https://github.com/widefield2sim/w2s}{the W2S github}. The structure is similar to the one of a Residual Dense Network but instead of the Residual Dense Blocks we have Residual Dense Networks.
\\
The input of our network is a one channel 64x64 pixels patch. The patch starts by going through a single convolutional layer, which is used to extract features, proceeds trough a series of RRDB blocks, get up-sampled by a factor of two and before exiting the network get down-sampled with maxPool of 2x2.
\subsection{RDDB blocks}
\begin{figure}[H]
    \centering
    \includegraphics[scale = 0.3]{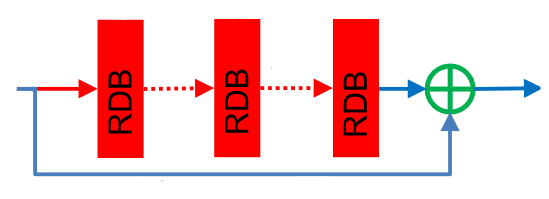}
    \caption{Picture of a single RRDB block}
    \label{fig:2}
\end{figure}

A RRDB block is composed by 3 RDB blocks in series. At the end of the 3 blocks the original input is added to the result [\ref{fig:2}]. The number of RRDB blocks can be modify at will. During our experimentations we tried different sizes; more specifically we ran the training with 4, 8 and 12 blocks. 
\subsection{RDB blocks}
Every residual dense block is composed by 5 convolutional layers. The first 4 operate a convolution with 64 filters, kernel size of 3x3 and zero-padding of 1. The activation function for those blocks is the Leaky-Relu with negative slope of 0.2. The input of each layer is the concatenation of all the outputs of the previous layers. The output of the fourth layer is then concatenated with the previous outputs and convolved with a single 64 filters layer of kernel size 3x3 and added to the initial input. 
\begin{figure}[H]
    \centering
    \includegraphics[scale = 0.3]{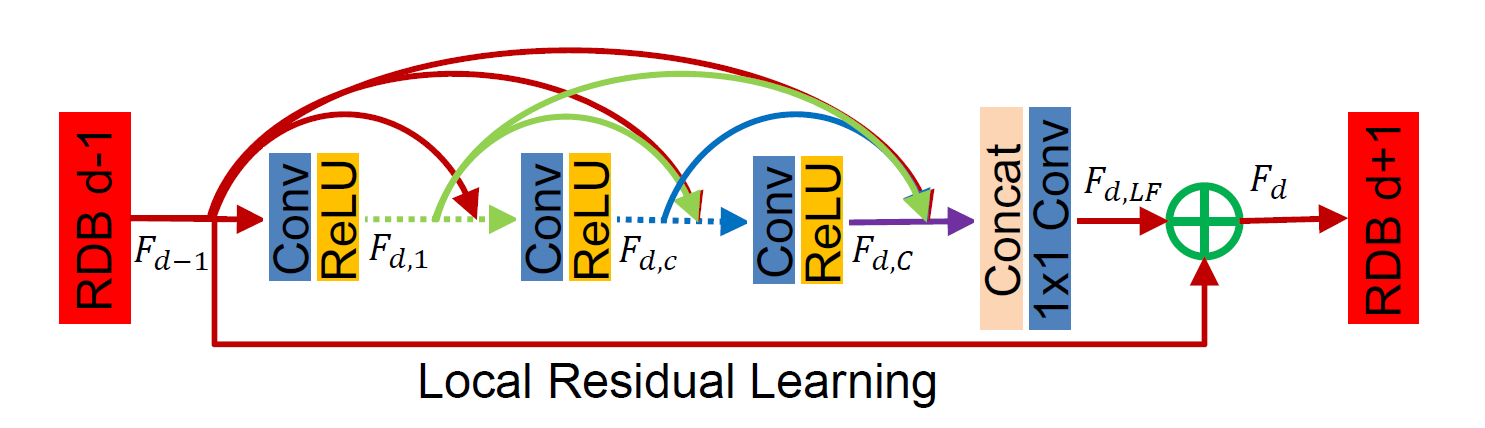}
    \caption{Example of a RDB block source: \cite{DBLP:journals/corr/abs-1802-08797}}
    \label{fig:3}
\end{figure}
\subsection{Post-processing}
We noticed that for most of the pictures exiting the network the intensity of the pixels was not covering the full possible spectrum. We thus decided to stretch the color histogram [\ref{eq:1}] in order to achieve better contrast and visibility of the cells.
\begin{equation} \label{eq:1}
g(x,y) = \frac{f(x,y)-fmin}{fmax-fmin}*255
\end{equation}
\begin{figure}
\begin{subfigure}{.5\columnwidth}
  \centering
  \includegraphics[scale = 0.2]{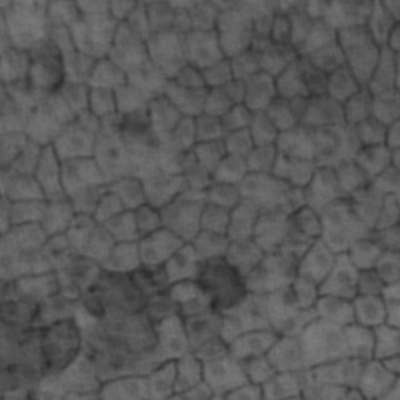}
  \caption{Before histogram stretching}
  \label{fig:sfig1}
\end{subfigure}%
\begin{subfigure}{.5\columnwidth}
  \centering
  \includegraphics[scale= 0.2]{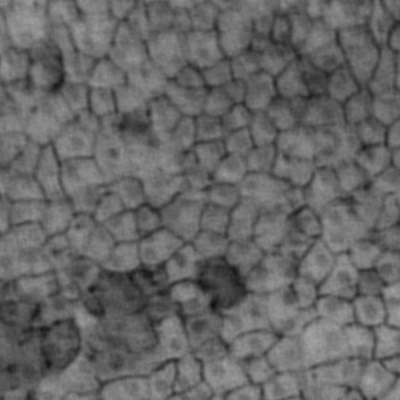}
  \caption{After histogram stretching}
  \label{fig:sfig2}
\end{subfigure}
\caption{Zoom-in of the output of the testing image 1\_2 from the network with training parameters 2 from [\ref{table 2}]}
\label{fig:fig4}
\end{figure}
\subsection{Training}
For the training, we experienced on modulation of different parameters: the number of RRDB blocs (i.e. depth of the network), the batch-size and learning rate. We train the network with the MSE loss function. In the beginning in addition to the our network we used a pretrained VGG feature extractor network, and trained with a texture loss that exploits the second-order statistics of feature maps \cite{arXiv:2003.05961}. But this slowed down the training and was not relevant enough for our solution, so we discarded it. The baseline batch-size was 16, we also tried different other sizes namely 32 and 64 when the cluster allowed us to, this accelerated the training and we increased the use of the GPU. In our different trainings the loss function was not converging, but had different values depending on the chosen parameters for the training as will be further shown in the results. We tried to modifiy the initial learning rate, since it can affect the gradient decent, to see if this would help. Decreasing it from $1e-4$ (the default value) to $1e-5$ and elevating it to $1e-3$, but this unfortunatly did not show significant changes except in elevating the loss. Overall we did about 10 different trainings, 1 epoch was lasting between a couple of hours to 10 hours when the cluster was overloaded. We trained with our different datasets using gaussian kernel and estimated kernels, we will present the most relevant solutions in the next section.
\section{Results}
We used \textit{peak signal to noise ration} and \textit{structural similarity} metrics to measure the quality of our test images. When comparing a degraded image with it's upgraded version, PSNR is an approximation to human perception of reconstruction quality. It is derived from the MSE. If we define I of size \textit{m}x\textit{n} as our sharp image and B its blurry version (or reconstructed version) the well known MSE is $$MSE=\frac{1}{mn}\sum_{m=0}^{m-1} \sum_{n=0}^{n-1}[ I(m,n) - B(m,n)]^2$$ and the derived PSNR is $$PSNR =10*log_{10}(\frac{MAX_{I}}{MSE}) $$ where $MAX_{I}$ is the maximum pixel value of the image I.\\
The SSIM index is based on perception-based model that considers image degradation as perceived change in structural information, versus MSE or PSNR that are based on the absolute errors. The SSIM index is calculated on various windows of an image. The measure between two windows x and y of common size NxN is.
\\
The following table describes the obtained values, and we will comment on them. We first show the PSNR and SSIM measures of the test images before feeding them to the network, the MSEs are in the order of 0.0008 for the first image and 0.001 for the second, this is because the second images has larger blurry parts than the second and this is also reflected by the lower metric results in general.
The image used for testing are images from the original data set aligned to the full stacks.
\begin{table}[H]
\caption{PSNR/SSIM metrics of the 8 Gaussian blurred testing images}
\centering
\resizebox{1.0\textwidth}{!}{\begin{tabular}{|c|c|c|c|c|c|c|c|c|}
\hline
      Image      & \textbf{1\_1} & \textbf{1\_2} & \textbf{1\_3} & \textbf{1\_4} & \textbf{2\_1} & \textbf{2\_2} & \textbf{2\_3} & \textbf{2\_4} \\ \hline
\textbf{PSNR \& SSIM} & 30.71/0.93    & 33.97/0.96    & 34.58/0.96    & 30.34/0.95    & 28.53/0.91    & 29.85/0.92    & 30.63/0.93    & 30.82/0.93    \\ \hline
\end{tabular}}
\label{table 1}
\end{table}

\begin{table}[H]
\caption{Training details}
\centering
\begin{tabular}{|c|c|c|c|c|}
\hline
\multicolumn{1}{|l|}{\textbf{Training}} & \textbf{Data set} & \multicolumn{1}{l|}{\textbf{Number of blocks}} & \multicolumn{1}{l|}{\textbf{Number epoch}} & \multicolumn{1}{l|}{\textbf{Validation MSE}}\\ \hline
\textbf{1} & Gaussian & 4 & 4 & 0.0017 \\ \hline
\textbf{2} & Gaussian & 4 & 9 & 0.0017 \\ \hline
\textbf{3} & Kernel estimate & 4 & 6 & 0.049 \\ \hline
\textbf{4} & Kernel estimate & 8 & 10 & 0.0117 \\ \hline
\end{tabular}
\label{table 2}
\end{table}

\begin{table}[H]
\caption{PSNR/SSIM metrics of the images [\ref{table 1}] after testing on trained nets [\ref{table 2}]}
\centering
\resizebox{1.0\textwidth}{!}{\begin{tabular}{|c|c|c|c|c|c|c|c|c|}
\hline
  \textbf{Image}          & \textbf{1\_1} & \textbf{1\_2} & \textbf{1\_3} & \textbf{1\_4} & \textbf{2\_1} & \textbf{2\_2} & \textbf{2\_3} & \textbf{2\_4} \\ \hline
\textbf{T1} & 30.51/0.92    & 33.45/0.95    & 34.14/0.96    & 30.13/0.92    & 28.52/0.90    & 29.80/0.92    & 30.56/0.93    & 30.67/0.93    \\ \hline
\textbf{T2} & 30.50/0.922   & 33.41/0.95    & 34.10/0.96    & 30.11/0.92    & 28.50/0.90    & 29.78/0.92    & 30.55/0.93    & 30.65/0.93    \\ \hline
\textbf{T3} & 16.66/0.82    & 16.72/0.84    & 16.69/0.84    & 16.62/0.83    & 17.70/0.82    & 17.79/0.83    & 17.80/0.83    & 17.81/0.83    \\ \hline
\textbf{T4} & 27.48/0.92    & 28.37/0.94    & 28.16/0.94    & 27.22/0.92    & 26.63/0.90    & 27.10/0.91    & 27.37/0.91    & 27.60/0.91    \\ \hline
\end{tabular}}
\end{table}
Looking at the results we see that we have a degradation in the PSNR and SSIM metrics, the main problem coming from the fact that our network had difficulties to learn from the data our loss stayed constant over the epochs (we did training until 20 epochs). We see that the training with gaussian blur gives the best results, with a pretty simplistic network, training with deeper networks with number of blocs of 12 or 16 gave values that were much worse with mean PSNR in the order of 20 dB and 10 dB respectively. But also a network with just one RRDB bloc and gaussian blur kernels gave bad results in the order of 10 dB in PSNR, so just going smaller is not the ultimate fix. For the estimated kernels the deeper net gave better results (comparison training 3 and 4). We here show comparison between training 4 and training 2 to demonstrate how small the changes are between epochs.  We believe that this comes from the lack of variety of the images since the dataset was essentially based on 10 images with were very close content. Also some blurs may have been too strong and too hard for the network to learn. In the end the goal of the project is to obtain images were the cell counting is facilitated, so this metrics may not that significant, and a cell counting algorithm would have been a better way to measure our improvements. Looking at the images we can see some slight enhancement, this gives hope that our solution with some changes could give better results. 
\begin{figure}
\begin{subfigure}{.5\columnwidth}
  \centering
  \includegraphics[scale = 0.32]{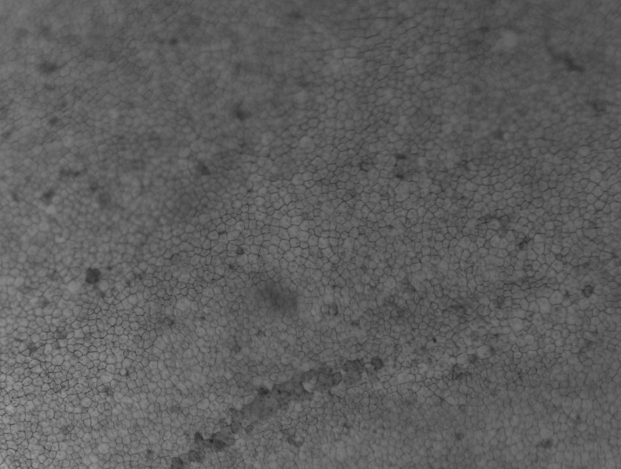}
  \caption{Input test image}
  \label{fig:sfig1}
\end{subfigure}%
\begin{subfigure}{.5\columnwidth}
  \centering
  \includegraphics[scale= 0.06]{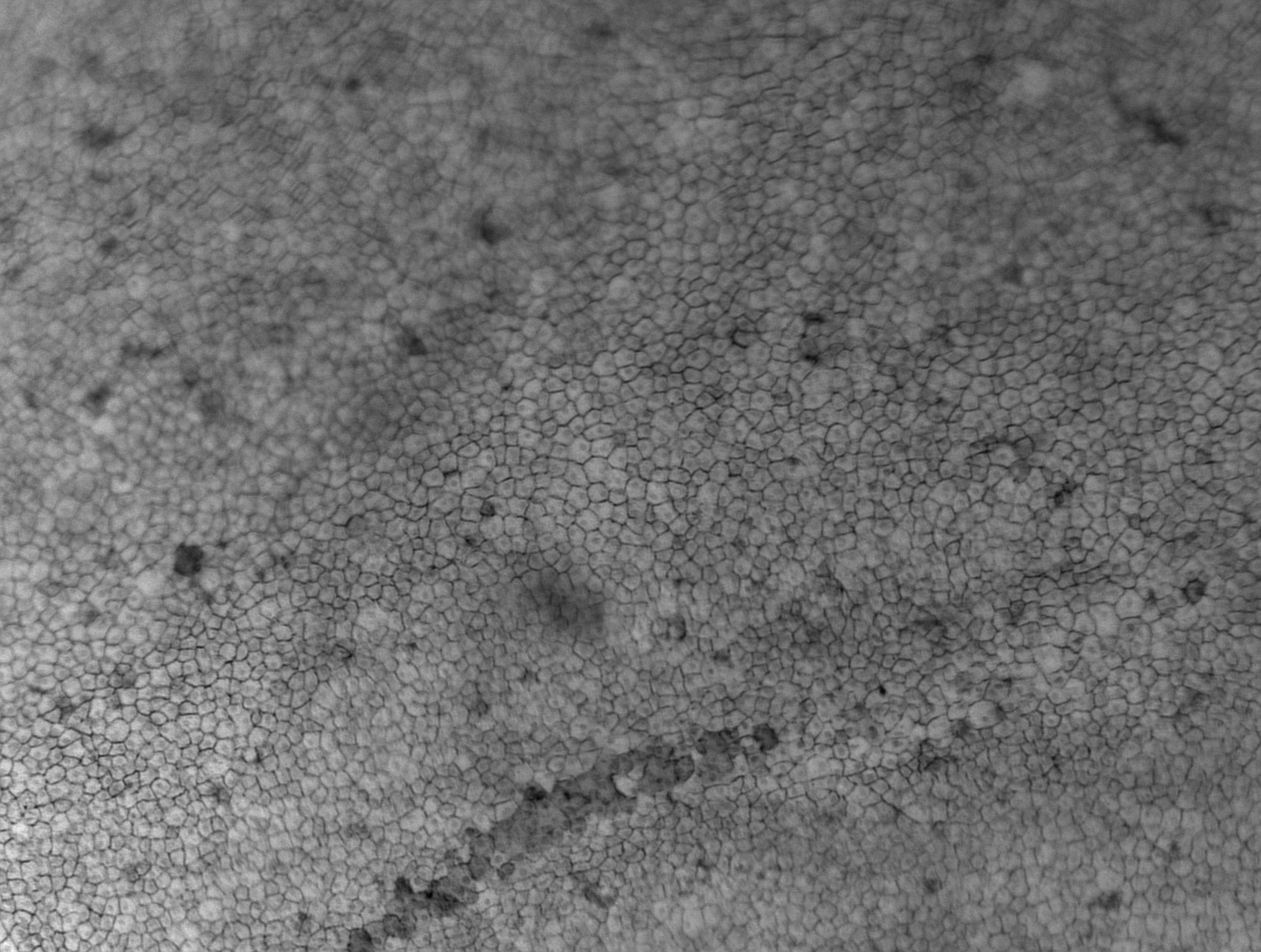}
  \caption{Output image post-processed}
  \label{fig:sfig2}
\end{subfigure}
\caption{Network test on the image 1\_2 generated with a Gaussian kernel with training parameters 2 from [\ref{table 2}]}
\label{fig:fig7}
\end{figure}
\subsection{Possible improvements}
Since our network does not converge we need to reflect on how to facilitate the learning process. A first fix is to increase and diversify the training dataset without loosing consistency. This could be done by adding other types of cell images captured with confocal microscopy. For example skin cell images could be close to eye cells and maybe less costly to obtain than cornea scans. The network architecture could be slightly modified as well, in a similar manner to the work by \cite{8963625} having first a down-scaling which makes the image sharper and then using SR network to keep the features of the down scaled layers in the upscaling process. Probably the learning process will have to be modified as well, as it is suggested in their work, by increasing the difficulty of the tasks to learn gradually. 
\section{Conclusion}
In this report we proposed an approach for Microscopy deblurring relying on a learning-based approach. Sharpness is already improved  with few training epochs despite the failure to train correctly the network. One positive point is the transition between patches, which is barely visible. This shows the advantages of deep learning over traditional methods where the difficulty is to handle more than one blur in one image. Having no discontinuities in the image could allow to better count the cells but we used usual metrics to compare our results (SSIM and PSNR). Applying the  counting algorithm currently used by ophtamologists would show how our method could impact diagnostics.

\section{Appendix}
\subsection{Code}
You can find our code on this \href{https://github.com/Plazzzzma/CP_project_2020}{GitHub}.
%
%

\bibliographystyle{alpha}
\bibliography{DLCMBD}
\end{document}